\documentclass[10pt,twocolumn,letterpaper]{article}

\usepackage[pagenumbers]{wacv}
\usepackage{cuted}
\usepackage{graphicx}
\usepackage{makecell}
\usepackage{booktabs}
\usepackage{float}
\usepackage{wasysym}
\usepackage{caption}
\usepackage{subcaption}
\usepackage{xcolor}
\definecolor{quan_red}{HTML}{D52728}
\definecolor{quan_orange}{HTML}{fE7F0E}
\definecolor{quan_blue}{HTML}{2177B4}
\newcommand\ignore[1]{}
\newcommand\mc{\multicolumn}

\ifdefined\hidecomments
  \newcommand\gal[1]{}
\else
   \newcommand\gal[1]{{\color{red}Gal: #1}}
\fi
\ifdefined\hidecomments
  \newcommand\ori[1]{}
\else
   \newcommand\ori[1]{{\color{blue}Ori: #1}}
\fi
\ifdefined\hidecomments
  \newcommand\dvir[1]{}
\else
   \newcommand\dvir[1]{{\color{orange}Dvir: #1}}
\fi

\definecolor{wacvblue}{rgb}{0.21,0.49,0.74}
\usepackage[pagebackref,breaklinks,colorlinks,allcolors=wacvblue]{hyperref}

\title{Per-Query Visual Concept Learning}
\author{
Ori Malca$^{1}$\quad
Dvir Samuel$^{1,2}$\quad
Gal Chechik$^{1,3}$\\
$^{1}$Bar‑Ilan University\quad
$^{2}$OriginAI\quad
$^{3}$NVIDIA\\\\[-0.5em]
\small\texttt{Project page: }\href{https://per-query-visual-concept-learning.github.io/}{\small\texttt{\textcolor[HTML]{ED018C}{https://per-query-visual-concept-learning.github.io}}}
}
\begin{document}
\maketitle
\begin{strip}
  \centering
  \includegraphics[width=1\textwidth]{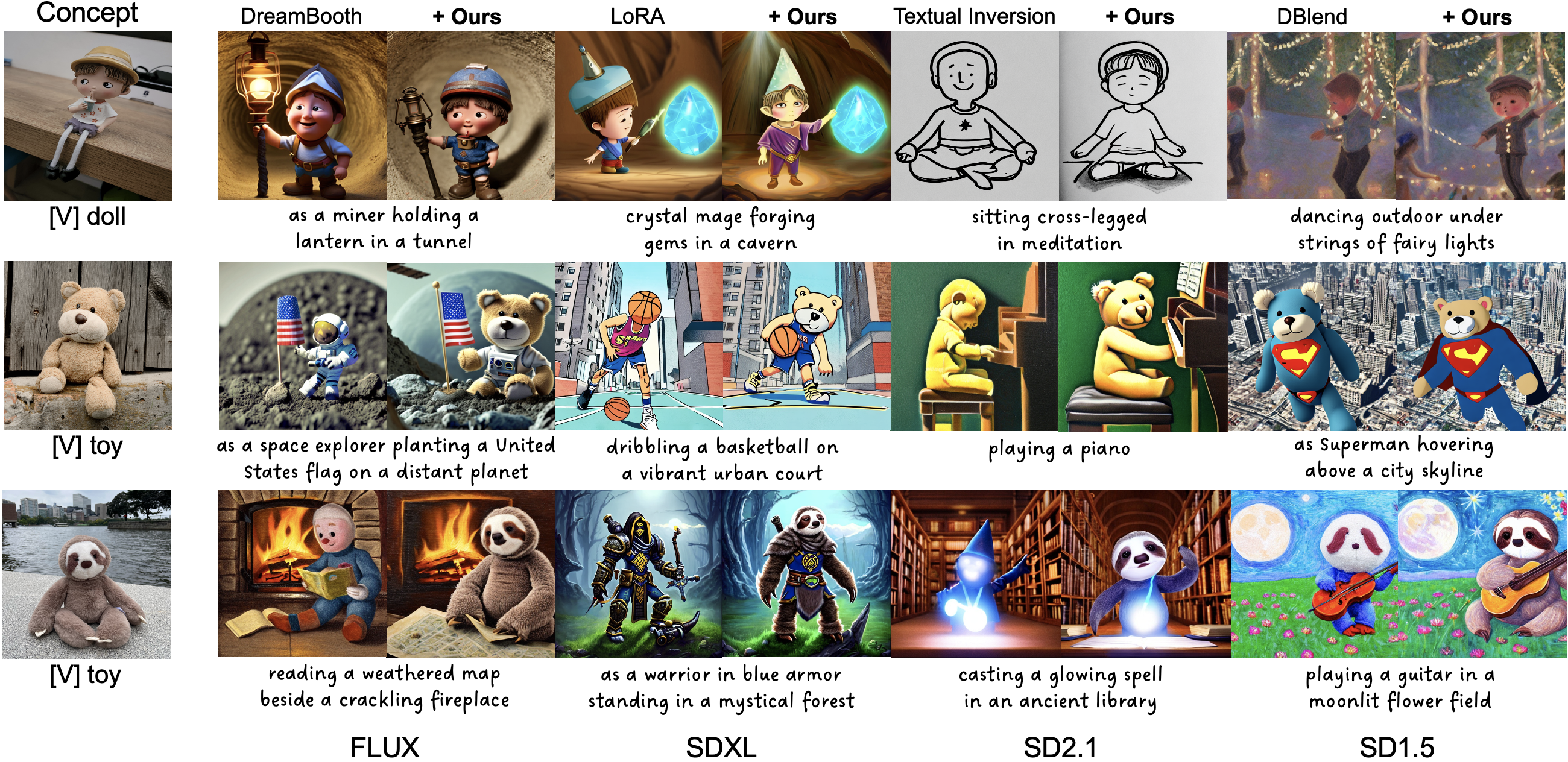}
  \captionof{figure}{Given a pretrained text-to-image personalization checkpoint, our method enhances per-query generation in terms of image alignment and text alignment, with just a single gradient update ($\sim$4 seconds on an NVIDIA H100 GPU). It is compatible with a wide range of personalization techniques (e.g., DreamBooth~\cite{ruiz2023dreambooth}, LoRA~\cite{LoRADiffusion}, Textual Inversion~\cite{gal2022image}, DBlend~\cite{ram2025dreamblend}) and supports various diffusion backbones, including UNet-based models (e.g., SDXL~\cite{podell2023sdxl}, SD~\cite{rombach2022high}) and transformer-based models (e.g., FLUX~\cite{labs2025flux1kontextflowmatching}).}
  \label{fig:general_results}
\end{strip}
\begin{abstract}
Visual concept learning, also known as Text-to-image personalization, is the process of teaching new concepts to a pretrained model. This has numerous applications from product placement to entertainment and personalized design. Here we show that many existing methods can be substantially augmented by adding a personalization step that is (1) specific to the prompt and noise seed, and (2) using two loss terms based on the self- and cross- attention, capturing the identity of the personalized concept. Specifically, we leverage PDM features - previously designed to capture identity - and show how they can be used to improve personalized semantic similarity. We evaluate the benefit that our method gains on top of six different personalization methods, and several base text-to-image models (both UNet- and DiT-based). We find significant improvements even over previous per-query personalization methods.
\end{abstract}

\section{Introduction}
\label{sec:intro}
Visual concept learning in generative text-to-image models is the problem of teaching a new concept to a pretrained model \cite{gal2022image,ruiz2023dreambooth, zhang2024survey}, usually from a handful of images of the target concept.  
This process, also known as \textit{personalization}, is typically achieved by tuning small parts of a pre-trained text-to-image model models, like the embedding layer of a token~\cite{gal2022image}, or a low-rank adaptation of the attention layers (DreamBooth LoRA~\cite{ruiz2023dreambooth, LoRADiffusion}). Personalization methods are evaluated in two main aspects: How well they capture the visual appearance of the target concept (\textit{image alignment}) and how well it can be controlled to agree with new text prompts (\textit{textual alignment}).  While significant progress has been made in this area, significant room is left for further improvement, especially when the number of target concept images that is available for fine-tuning is small. 

Here we propose to improve generation of a new visual concept using two observations. First, that the (reconstruction) losses used for personalization were typically not designed for capturing concept identity.  Second, that personalization methods typically fine-tune models  using a diverse set of prompts and seeds, with the goal of using the learned concept with any seed and prompt at inference time (see \cite{arar2024palp,agarwal2025alignit}). To address this, we introduce a novel loss function tailored to better preserve visual identity. Our approach is inspired by recent work \cite{samuel2024s}, which found that object appearance is largely encoded in specific layers of the self-attention mechanism. These features were originally used for instance retrieval and segmentation, and here we adapt this insight to improve the training process of existing personalization methods. To further improve text-image alignment, we show that fine-tuning with a fixed prompt and seed significantly enhances generation quality.

We describe an "add-on" fine tuning procedure that can be used on top of existing methods. For instance, once a visual concept has been learned using some method, we perform feature extraction to compute identity-sensitive features from the self-attention part, and prompt-sensitive features from the cross-attention part, and update the model using a single gradient update.  We find that this single update step provides significant gains in terms of visual alignment and text alignment. The procedure is simple, fast, and easy to implement. We tested it on top of six different personalization methods and backbone text-to-image diffusion models, (both SD~\cite{rombach2022high}, SDXL~\cite{podell2023sdxl} and Flux~\cite{labs2025flux1kontextflowmatching}) and found that it provides consistent gains.

The paper therefore makes the following contributions: 
\textbf{(1)} An identity-preserving loss based on self and cross attention. \textbf{(2)} New SoTA results in personalization using a single image of the target concept. \textbf{(3)} It demonstrates the benefit of fine-tuning with a specific seed and prompt.

\begin{figure}[h]
    \centering
    \includegraphics[width=1\linewidth]{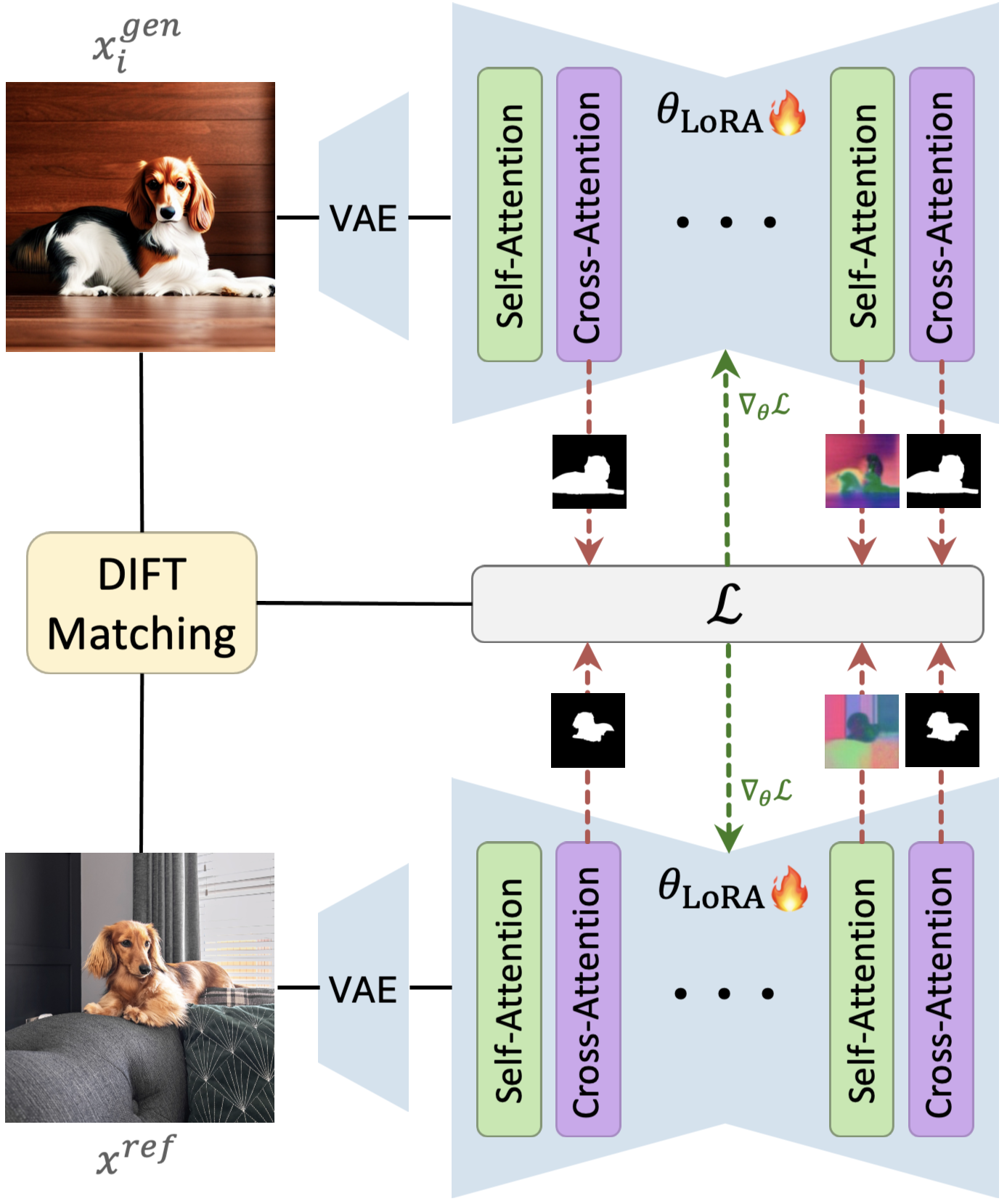}
    \caption{\textbf{Method Overview.} We enhance personalized text-to-image models by computing self- and cross-attention features from a single denoising step of a generated image $x^{\text{gen}}$ and a reference image $x^{\text{ref}}$. Using DIFT, we match these features and define losses $\mathcal{L}_{\text{SA}}$, $\mathcal{L}_{\text{CA}}$, and $\mathcal{L}_{\text{LDM}}$, which are combined to update the personalization tuning parameters via gradient descent.}
    \label{fig:method}
\end{figure}

\section{Related work}
\label{sec:related_work}
Text-to-image diffusion models are trained to generate samples from a conditional data distribution by progressively denoising a variable initially sampled from a Gaussian distribution. Recent advances in text-to-image synthesis have been driven by large-scale models trained on massive web-sourced datasets \cite{rombach2022high, saharia2022photorealistic, ramesh2021zero, sun2023emu, ramesh2022hierarchical, yu2022scaling}.

Personalization in generative text-to-image models involves teaching a pre-trained model a new concept using only a few images. Early approaches tackled this problem by optimizing word embeddings \cite{gal2022image, voynov2023p+, alaluf2023neural, wei2023elite} or fine-tuning model weights \cite{ruiz2023dreambooth}. While fine-tuning typically yields better subject fidelity and photorealism—thanks to its higher expressive capacity—it tends to compromise prompt fidelity and image diversity as training progresses. Subsequent methods have aimed to mitigate this trade-off.

CustomDiffusion~\cite{kumari2023multi} constrains updates to the text-image cross-attention layers and regularizes training using retrieved class-specific images. Perfusion~\cite{tewel2023key}, inspired by ROME~\cite{Meng2022LocatingAE}, applies a gated rank-one update to the key and value projection matrices, anchoring the key space within the subject’s super-category to curb overfitting. DBlend~\cite{ram2025dreamblend} guides generations from an overfit checkpoint by matching the cross-attention maps of an underfit one, adjusting noise levels at each diffusion step accordingly. AttnDB~\cite{pang2024attndreambooth} proposes a multi-stage training scheme that learns embedding alignment, attention maps, and subject identity separately, enhancing both identity preservation and prompt fidelity.

Another line of research explores encoder-based methods \cite{shi2024instantbooth, gal2023encoder, jia2023taming, chen2024subject, ye2023ip, wang2024instantid, ma2023unified}, which avoid subject-specific fine-tuning and model-specific weight storage. While some of these methods \cite{gal2023encoder, jia2023taming, shi2024instantbooth, wang2024instantid} are domain-specific (e.g., human faces or pets) and may still involve limited fine-tuning, others \cite{li2024blip, chen2024subject} strive for more general-purpose, zero-shot personalization. However, such models often depend on large-scale pretraining involving diffusion models in the loop. In some cases \cite{chen2024subject}, they even rely on supervision from expert-personalized models to construct their training datasets.

In contrast to these encoder-based or architectural modifications, our approach remains firmly grounded in fine-tuning. We leverage the final stages of training to extract and synthesize high-quality images enriched with subject identity cues. These synthesized images are then used to train more efficient personalization models for inference-time deployment. Rather than discarding fine-tuning due to its computational cost, we embrace it as both a powerful personalization mechanism and a source of high-fidelity supervision. Our method is therefore positioned within the fine-tuning paradigm, but repurposed it to enhance prompt adherence and subject alignment while enabling efficient downstream inference—offering a unique balance between quality, adaptability, and efficiency.

\section{Method}
\label{sec:method}

\subsection{Revisiting the Loss Functions Used in Current Personalization Pipelines}
\label{sec:std_loss}

Most personalization methods, whether they introduce a new token in the text encoder
(\emph{textual inversion}~\cite{gal2022image}), adapt a small set of weights with low-rank
updates (\emph{DreamBooth LoRA}~\cite{ruiz2023dreambooth, LoRADiffusion}), or fine-tune a lightweight subset
of layers (\emph{CustomDiffusion}~\cite{kumari2023multi}), often optimize \emph{exactly}
the same objective used to train the original diffusion backbone.
Given an image $x_{0}$ paired with a text prompt $\tau$, Gaussian noise
$\epsilon\!\sim\!\mathcal{N}(0,\mathbf{I})$ is added according to the forward process,
yielding
$$
  x_{t}
  =\sqrt{\bar\alpha_{t}}\;x_{0}
  +\sqrt{1-\bar\alpha_{t}}\;\epsilon
$$
The model parameters $\theta$ are then updated by minimizing the noise-prediction loss
\begin{equation*}
  \mathcal{L}_{\text{diff}}=
  \mathbb{E}_{t,\epsilon,z_{0},\tau}\!\bigl[
    \,\| \epsilon-
      \epsilon_{\theta}(z_{t},t,\tau) \|_{2}^{2}
  \bigr]
\end{equation*}
This objective treats all pixels equally, regardless of whether they correspond to the object of interest, background, or unrelated regions. Moreover, since it uses a per-pixel $L_2$ loss, the model is not explicitly encouraged to reconstruct the visual identity of the object; it may converge to solution with low-loss values that perceptually deviate from the reference concept identity.

To address these limitations, we propose two key enhancements: (1) Object-Focused Feature Alignment: We first localize the object in both the reference and generated images, then extract features specifically from those regions. This enables us to focus the training signal on the object itself, rather than the entire scene. (2) Appearance and Prompt Conditioning: The extracted features disentangle into appearance features (capturing the visual identity) and prompt features (capturing semantic alignment). By supervising the model with both, we explicitly enforce appearance consistency with the reference image while ensuring fidelity to the provided prompt.

\subsection{Our approach}
\newcommand\zgt{z_{T}}
\newcommand\xr{x^{\text{ref}}}
\newcommand\zr{z_{0}^{\text{ref}}}
\newcommand\Fr{F_{0}^{\text{ref}}}
\newcommand\Mr{M_{0}^{\text{ref}}}

\newcommand\zg{z_{0}^{\text{gen}}}
\newcommand\Fg{F_{0}^{\text{gen}}}
\newcommand\Mg{M_{0}^{\text{gen}}}

\newcommand\pr{p_^{ref}}

Our method is designed to be an easy-to-use add-on on top of existing personalization methods like DreamBooth~\cite{ruiz2023dreambooth} or Textual-Inversion~\cite{gal2022image}. It can be applied to a range of base methods and models, from Unet-based diffusion models (e.g SD~\cite{rombach2022high} and SDXL~\cite{podell2023sdxl}) to DiT-based flow models~\cite{lipman2022flow} (e.g FLUX~\cite{labs2025flux1kontextflowmatching} and SD3~\cite{esser2024scaling}). See Figure \ref{fig:method} for an overview of our approach.

In short, our method has two key ideas. First, we start with a specific prompt a specific noise seed. Second, we compute two losses based on the self- and cross- attention features, using a single denoising step of the clean $z_0$, and use DIFT~\cite{tang2023DIFT} to match features between the generated and reference image.   

To describe in more detail, we define the following notation. Assume we have a pre-trained (personalized) text-to-image model with tunable parameters $\theta$, and we are given a target prompt $p$ a noise seed $\zgt$, and a reference image $\xr$ of a personalized concept to be learned. The goal is to generate an image based on the prompt that contains the personalized concept.
As commonly done, we assume that the target prompt $p$ has a format like "A photo of a [V] [dog] in the park" where [V] denotes the target concept in the prompt. We also assume that the model was already trained with a reference prompt of the form "a photo of [V] [dog]", using some base personalization method like DB-LoRA~\cite{ruiz2023dreambooth, LoRADiffusion}, Textual-Inversion~\cite{gal2022image}, or others~\cite{ram2025dreamblend, arar2024palp, LoRADiffusion}.  

\noindent Our method follows four simple steps: 

\paragraph{(A) Compute a representation for the reference image:} 
We compute the latent $\zr = \text{VAE}(\xr)$, then perform a single denoising step of $\zr$, with a denoising parameter $\alpha$ determined with a parameter $t=1000$. This denoising step is solely for feature extraction and does not change the latent. We then compute two representations of the reference image: (1) appearance features $\Fr$ of the latent $\zr$ from self-attention layers, following the recent PDM method \cite{samuel2024s}; and (2)  cross-attention maps for the concept [V], denoted by $\Mr$. For example, in models based on UNet, we took attention maps from all 16 layers of the UNet in Stable Diffusion. 

\paragraph{(B) Compute a representation for a generated image:} This follows the same steps like those for reference image. Namely, we first generate a $\zg$ using the pretrained model with the prompt $ p $. Then, compute a self attention $\Fg$ and cross attention features $\Mg$ for the generated latent $\zg$. 

\paragraph{(C) Compute DIFT matching between the two latents $\zr, \zg$:} It gives us a list of $n$ matched pairs of points $(\zr(i), \zg(i))$, $i=1,\ldots,n$. 

\paragraph{(D) Compute three loss components:} 
Using the features extracted as descried above, we compute three losses. One based on self-attention, one based on cross-attention, and based on the standard reconstruction loss. 
\begin{equation}
    \mathcal{L}_{\mathrm{SA}} = \| \Fr - \Fg \|_{_{DIFT}}
\end{equation}
\begin{equation}
    \mathcal{L}_{\mathrm{CA}} = \| \Mr - \Mg\|_{_{DIFT}}
\end{equation}
\begin{equation}
    \mathcal{L}_{\mathrm{LDM}} =
    \mathbb{E}_{
        \zr \sim \mathcal{E}(\xr), p,\epsilon \sim \mathcal{N}(0,1), t}
    \!\Big[\|\epsilon - \epsilon_{\theta} (z^{\text{ref}}_{t},t,p) \|_{2}^{2}\Big]
\label{eq:LDM_Loss}
\end{equation}
where the norm $\| \cdot \|_{DIFT}$ is an $L_2$ norm taken over the set of matching point pairs computed using DIFT in step C.    

Combining all losses gives:
\begin{equation}
    \mathcal{L}  =
    \lambda_{\text{SA}} \mathcal{L}_{\text{SA}}    +
    \lambda_{\text{CA}} \mathcal{L}_{\text{CA}}    +
    \lambda_{\text{LDM}} \mathcal{L}_{\text{LDM}} \quad
\end{equation}
where all $\lambda$ values are tradeoff coefficients treated as hyper parameters. 

Finally, we update the parameters of the personalized model $\theta$ using a gradient step over $d\mathcal{L} / d\theta$. The process can be repeated for several gradient steps, but in practice, we found that a single gradient step provided significant improvement.

\paragraph{Richer representations}
The method described above use only features computed from $\zr$ and $\zg$, the cleanest latent. We also explored the case where features are computed from multiple different cleaning steps. For the reference image, this requires inverting the image into $z_T^{ref}$, and selecting $T$ different denoising steps along the denoising path. For the generated image, we similarly take the corresponding denoising steps along the path. Figure~\ref{fig:ts_time_opt1_ablation} below shows that using more features does improve generation quality, but even $T=1$ provides a significant benefit.
\begin{figure*}[!h]
  \centering
  \includegraphics[width=1\linewidth]{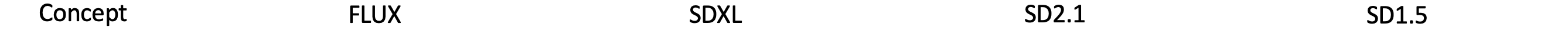}
  \includegraphics[width=1\linewidth]{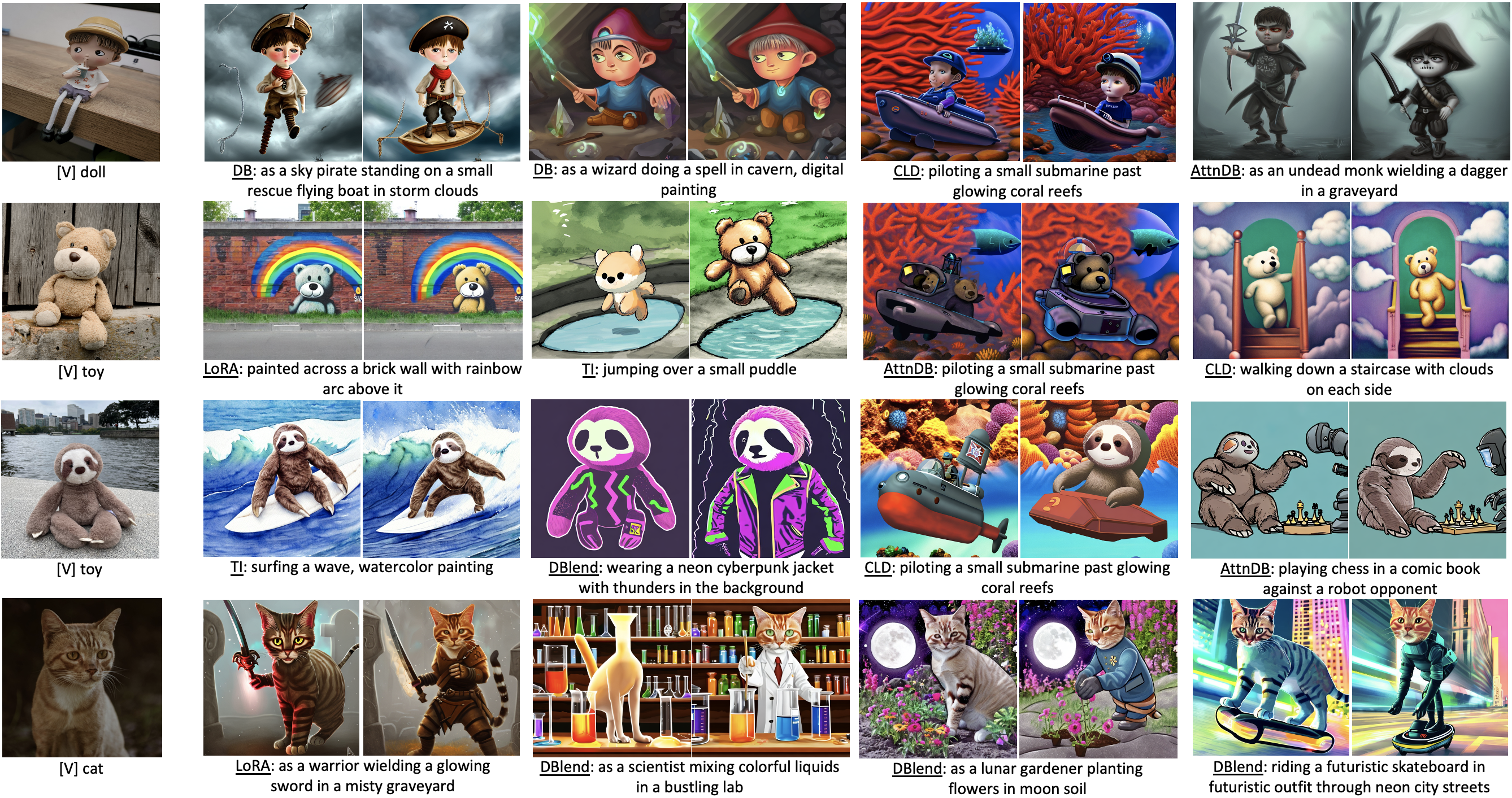}
  \includegraphics[width=1\linewidth]{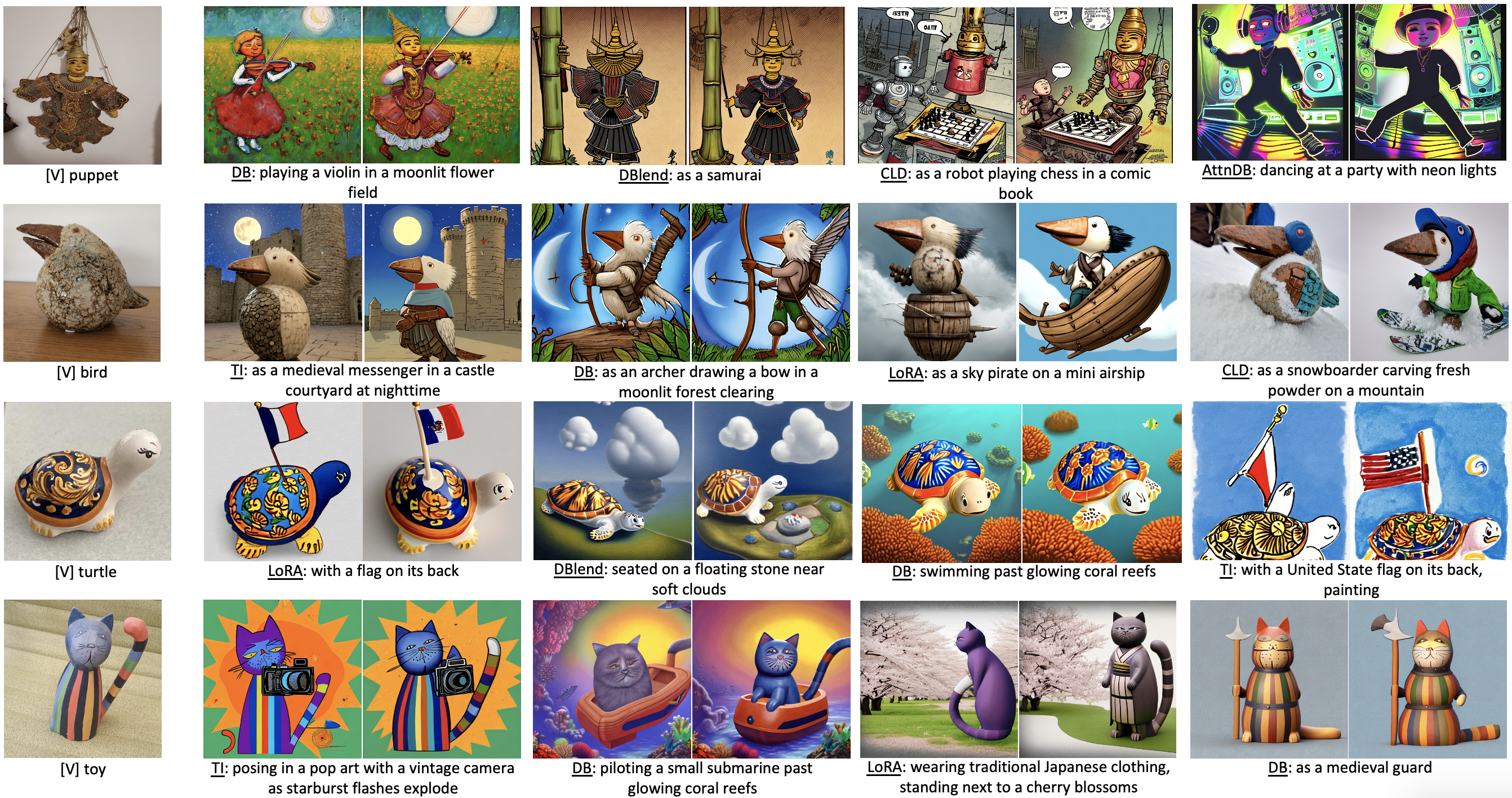}
  \caption{\textbf{Qualitative comparison}. We present images generated by various personalization methods without our method and with it, including DB~\cite{ruiz2023dreambooth}, TI~\cite{gal2022image}, DBlend~\cite{ram2025dreamblend}, CLD~\cite{huang2024classdiffusion}, AttnDB~\cite{pang2024attndreambooth}, and LoRA~\cite{LoRADiffusion}, across 4 different backbones (FLUX~\cite{labs2025flux1kontextflowmatching}, SDXL~\cite{podell2023sdxl}, SD2.1, SD1.5~\cite{rombach2022high}). The left column shows one example of the target concept, and then every pair of columns shows the generated images with and without our method. Each pair uses the same prompt and seed. Adding our method to those personalization approaches yields better performance in text alignment and identity preservation compared to these baselines. Prompts were generated using ChatGPT.}
  \label{fig:quaL_vs_SOTA}
\end{figure*}

\section{Experiments}
\label{sec:experiments}
This section begins with a description of our implementation setup. We then assess our approach on commonly used benchmarks and show it improves all baselines across different text-to-image diffusion models. Finally, we carry out ablation studies to examine the individual impact of each component in our method.

\subsection{Implementation and Evaluation Setup}
\subsubsection{Implementation Details}
Our implementation is based on the publicly available Stable Diffusion V1.5, V2.1~\cite{rombach2022high}, SDXL~\cite{podell2023sdxl}, and FLUX~\cite{labs2025flux1kontextflowmatching} architectures. All experiments are conducted on a single NVIDIA H100 GPU. Both $\lambda_{\text{SA}}$ and $\lambda_{\text{CA}}=1$ are set to 1. The optimization process of our method takes ~4 seconds to tune a personalized model for a specific query. We perform a single optimization step of our method with a batch size of 1 and the same learning rate used during the training procedure of the personalized model.

For the first stage of personalization training (used as a baseline upon which our method builds), we adopted the authors' original implementations and hyperparameters whenever available. While most methods provided public code, we were unable to obtain implementations for DBlend~\cite{ram2025dreamblend}, AlignIT~\cite{agarwal2025alignit}, and PALP~\cite{arar2024palp}. For these, we re-implemented the algorithms based on the descriptions in their respective papers.

\subsubsection{Evaluation Setup}
We evaluate our method in two sets of experiments:

\textbf{(A)} On top of six personalization method, including DreamBooth~\cite{ruiz2023dreambooth}, LoRA~\cite{LoRADiffusion}, Textual Invresion~\cite{gal2022image}, DBlend~\cite{ram2025dreamblend}, AttnDB~\cite{pang2024attndreambooth}, and CLD~\cite{huang2024classdiffusion}. Results are provided with (w/) and without (w/o) our method across three diffusion models: Stable Diffusion (SD)~\cite{rombach2022high}, Stable Diffusion XL (SDXL)~\cite{podell2023sdxl}, and FLUX~\cite{labs2025flux1kontextflowmatching}. For the quantitative evaluation, we follow the literature and evaluate our approach on the DreamBooth's dataset.

\textbf{(B)} Against per-query methods, including AlignIT~\cite{agarwal2025alignit}, and PALP~\cite{arar2024palp}, on top of different personalization baselines, including Textual Inversion~\cite{gal2022image}, LoRA~\cite{LoRADiffusion}, DB~\cite{ruiz2023dreambooth}, with FLUX~\cite{labs2025flux1kontextflowmatching}, SDXL~\cite{podell2023sdxl}, and SD2.1~\cite{rombach2022high} as backbones. For the quantitative evaluation, we use the TI and CustomConcept101 datasets proposed by \cite{gal2022image} and \cite{choi2023custom}, respectively.
\subsection{Results}
Since our method can be applied on top of existing personalization approaches, we evaluate its benefits by comparing images generated with and without our method.

Specifically, we conducted qualitative and quantitative evaluations between personalization methods \cite{pang2024attndreambooth, kumari2023multi, huang2024classdiffusion, ram2025dreamblend, gal2022image, ruiz2023dreambooth} with (w/) and without (w/o) our method across FLUX~\cite{labs2025flux1kontextflowmatching}, SDXL~\cite{podell2023sdxl}, SD1.5 and SD2.1~\cite{rombach2022high} backbones. We also provide ablation studies to explore the impact of the components of our method.

\subsubsection{Qualitative Evaluation}
For evaluation, we used ChatGPT to generate complex prompts that satisfy the following criteria: a prompt was required to contain several elements such as appearance (“as a wizard”), action ("doing a spell"), scene (“in a cavern”), and style (“digital painting”).

\begin{figure*}[!h]
  \centering
  \includegraphics[width=1\linewidth]{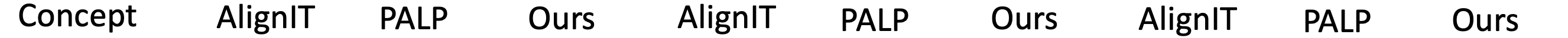}
  \includegraphics[width=1\linewidth]{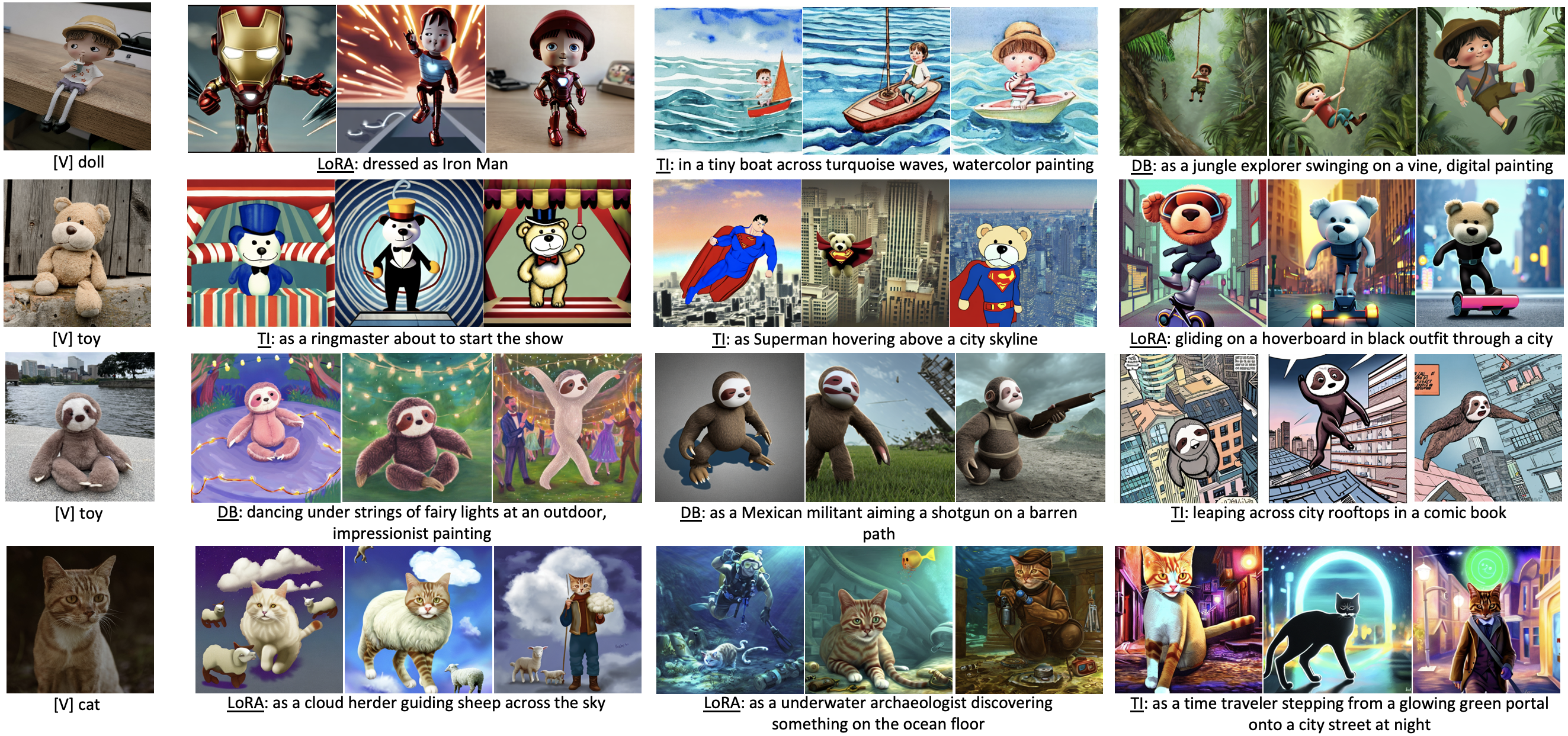}
  \includegraphics[width=1\linewidth]{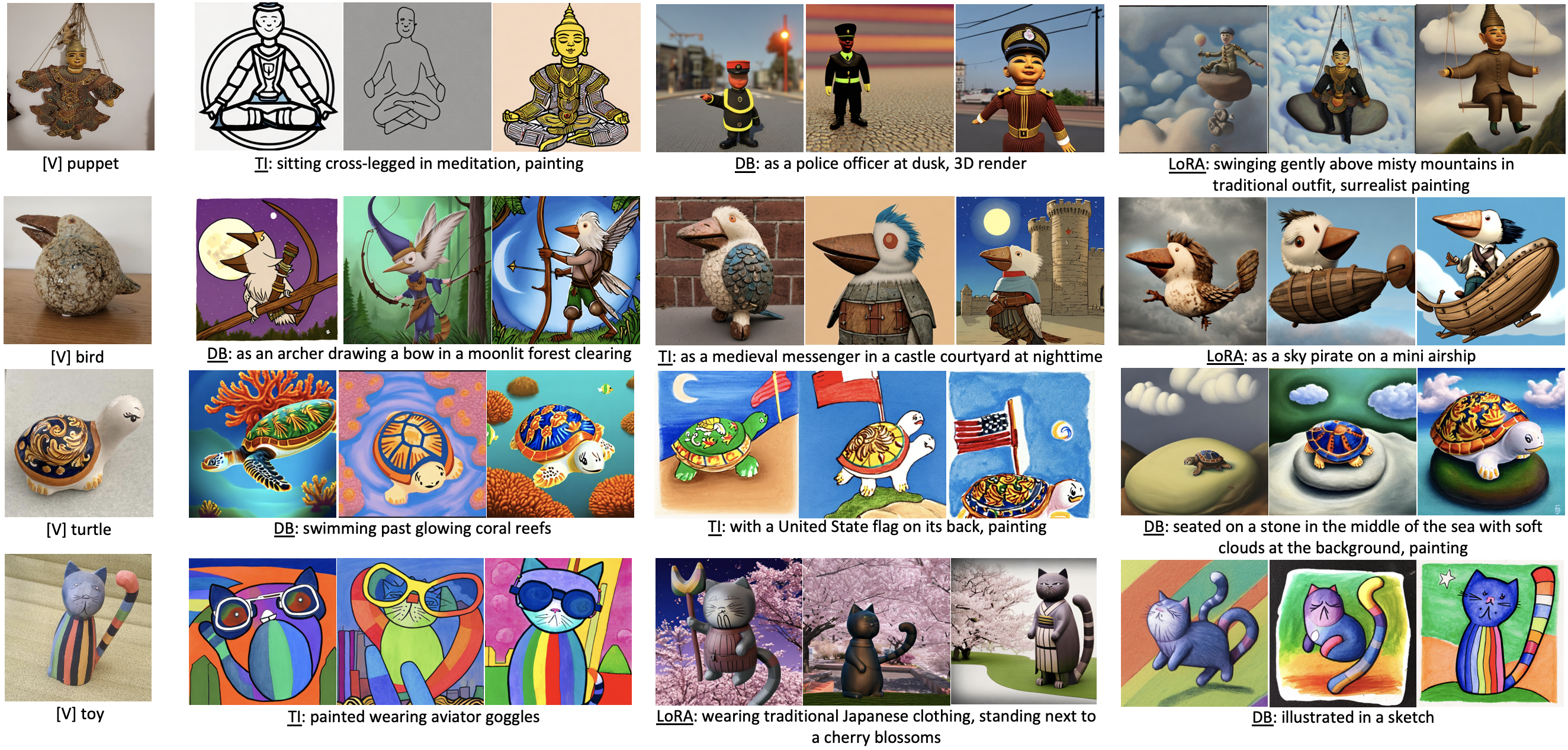}
  \includegraphics[width=1\linewidth]{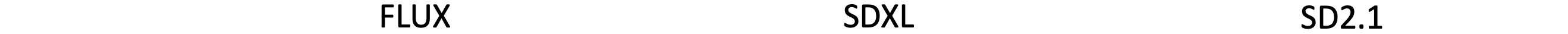}
  \caption{\textbf{Qualitative comparison}. We present images generated by 3 personalization methods (including DB~\cite{ruiz2023dreambooth}, TI~\cite{gal2022image}, LoRA~\cite{LoRADiffusion}) using 3 per-query methods (including AlignIT~\cite{agarwal2025alignit}, PALP~\cite{arar2024palp}, Ours), across 3 different backbones (FLUX~\cite{labs2025flux1kontextflowmatching}, SDXL~\cite{podell2023sdxl}, SD2.1~\cite{rombach2022high}). Our method demonstrates superior performance in text alignment and identity preservation compared to these per-query methods. Our per-query method, is the only method that successfully generates identity preserved and text-aligned personalized images for these complex queries.}
  \label{fig:quaL_vs_PQ}
\end{figure*}

Figure~\ref{fig:quaL_vs_SOTA} provides a visual comparison of images generated with various state-of-the-art text-to-image personalization methods with (w/) and without (w/o) our method. As observed, for certain queries, without our method, these baselines either: (1) overlooks the learned concept, producing images that solely reflect other prompt tokens (e.g., "puppet as a robot ..."), (2) tends to overfit the new concept, failing to compose it in novel scenes or actions (e.g., "toy surfing ..."), (3) struggle to achieve text-aligned generations (e.g., "doll as a sky pirate ..."), (4) fail to preserve the identity of the learned concept (e.g., "puppet playing a violin ..."). Incorporating our method provides successful text-aligned and identity-preserved personalization for these complex queries.
Figure~\ref{fig:quaL_vs_PQ} provides a visual comparison between images generated with our method and with two prominent per-query methods (AlignIT~\cite{agarwal2025alignit} and PALP~\cite{arar2024palp}) across different baselines (including DB~\cite{ruiz2023dreambooth}, LoRA~\cite{LoRADiffusion}, TI~\cite{gal2022image}). Similarly, we observed (1) "puppet sitting ...", (2) "toy dancing ...", (3) "as a Mexican militant ...", (4) "toy as a ringmaster ...", respectively demonstrating the same 4 failure modes noted in the preceding paragraph. Our per-query method, is the only method that successfully generates identity preserved and text-aligned personalized images for these complex prompts. Additional qualitative results can be found in Figures~\ref{fig:qlc_vs_SOTA_more} and \ref{fig:qlc_vs_PQ_more}.

\begin{figure}[t]
    \centering
    \includegraphics[width=0.88\linewidth]{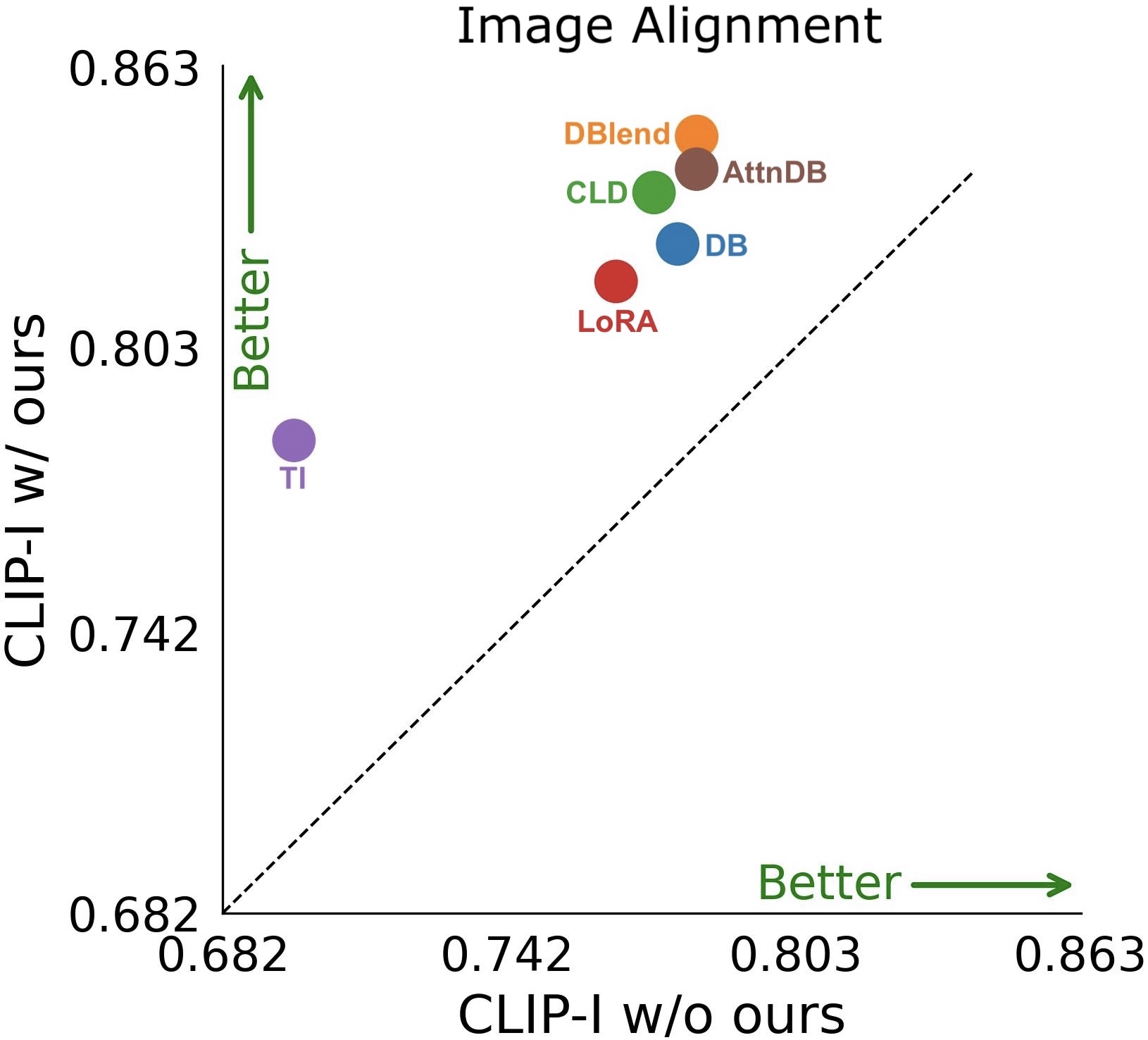}
    \newline~\newline
    \includegraphics[width=0.88\linewidth]{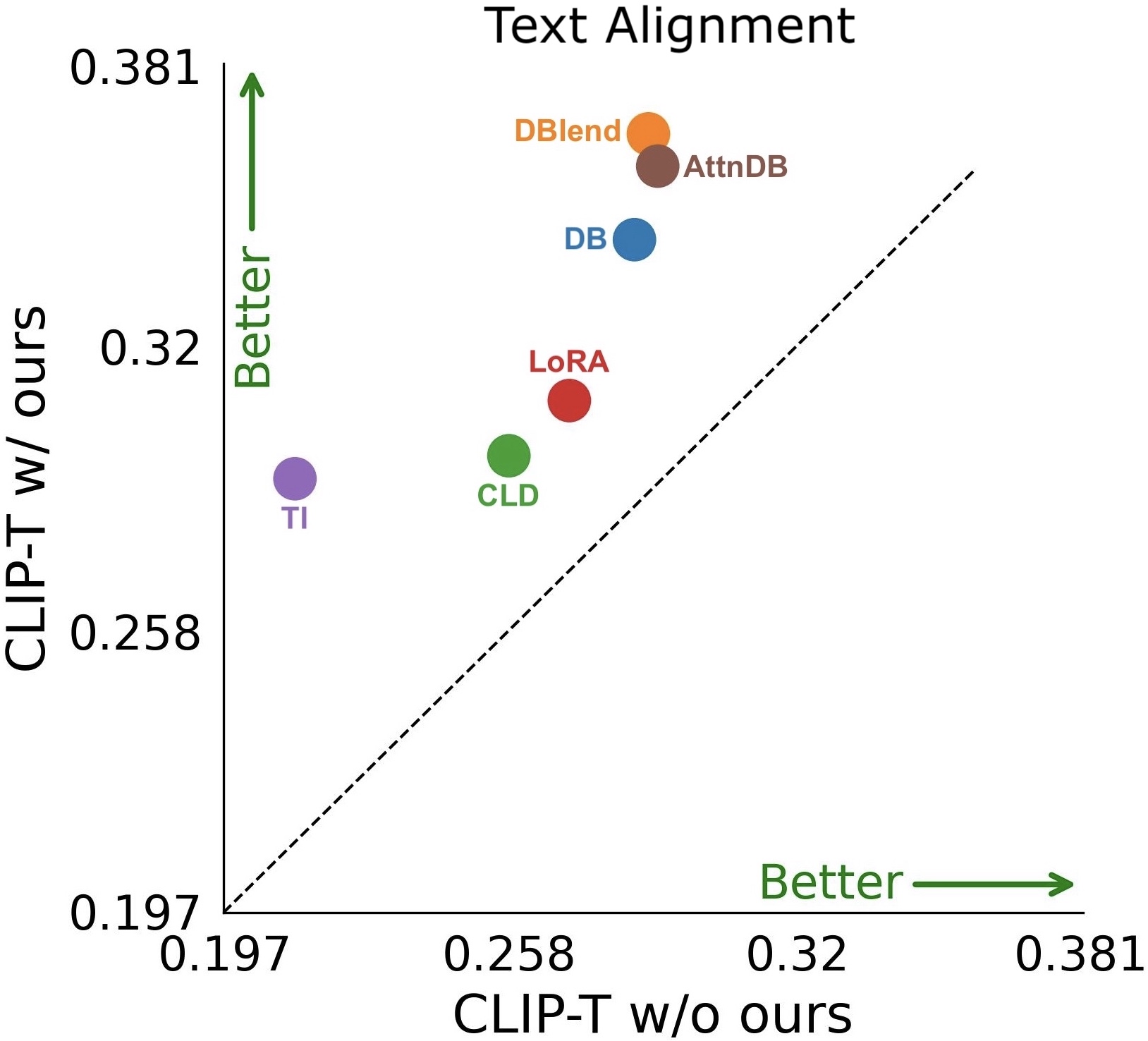}
    \caption{Effect of adding our method to baseline personalization models. shown are image alignment (top) and text alignment (bottom) of several personalization approaches with (w/) and without (w/o) the integration of our method.}
    \label{fig:qnc_vs_SOTA}
    \vspace{-2pt}
\end{figure}

\begin{figure}[h]
    \centering
    \includegraphics[width=0.68\linewidth]{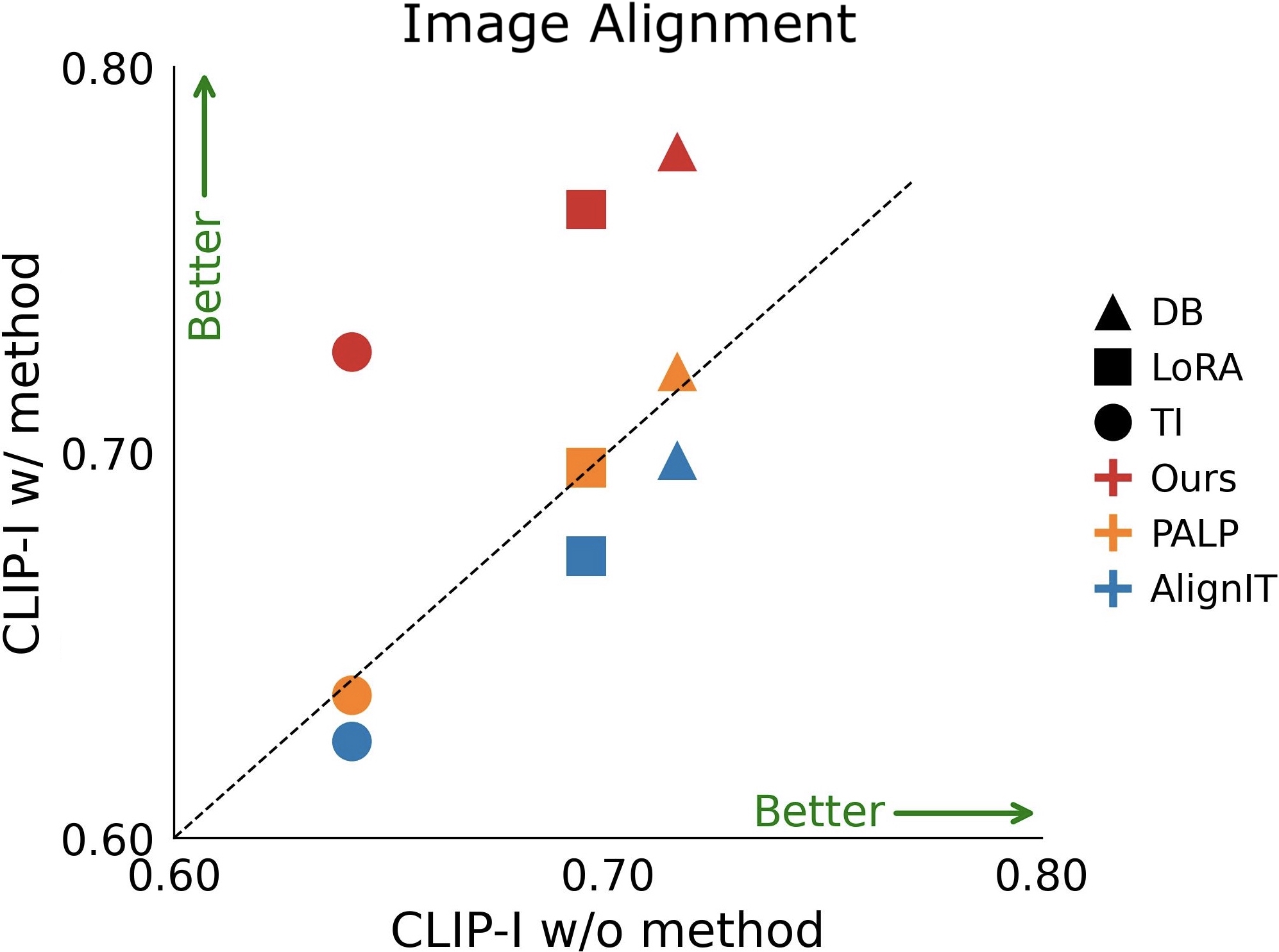}
    \includegraphics[width=0.68\linewidth]{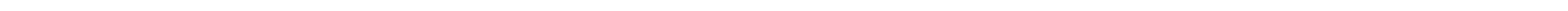}
    \includegraphics[width=0.68\linewidth]{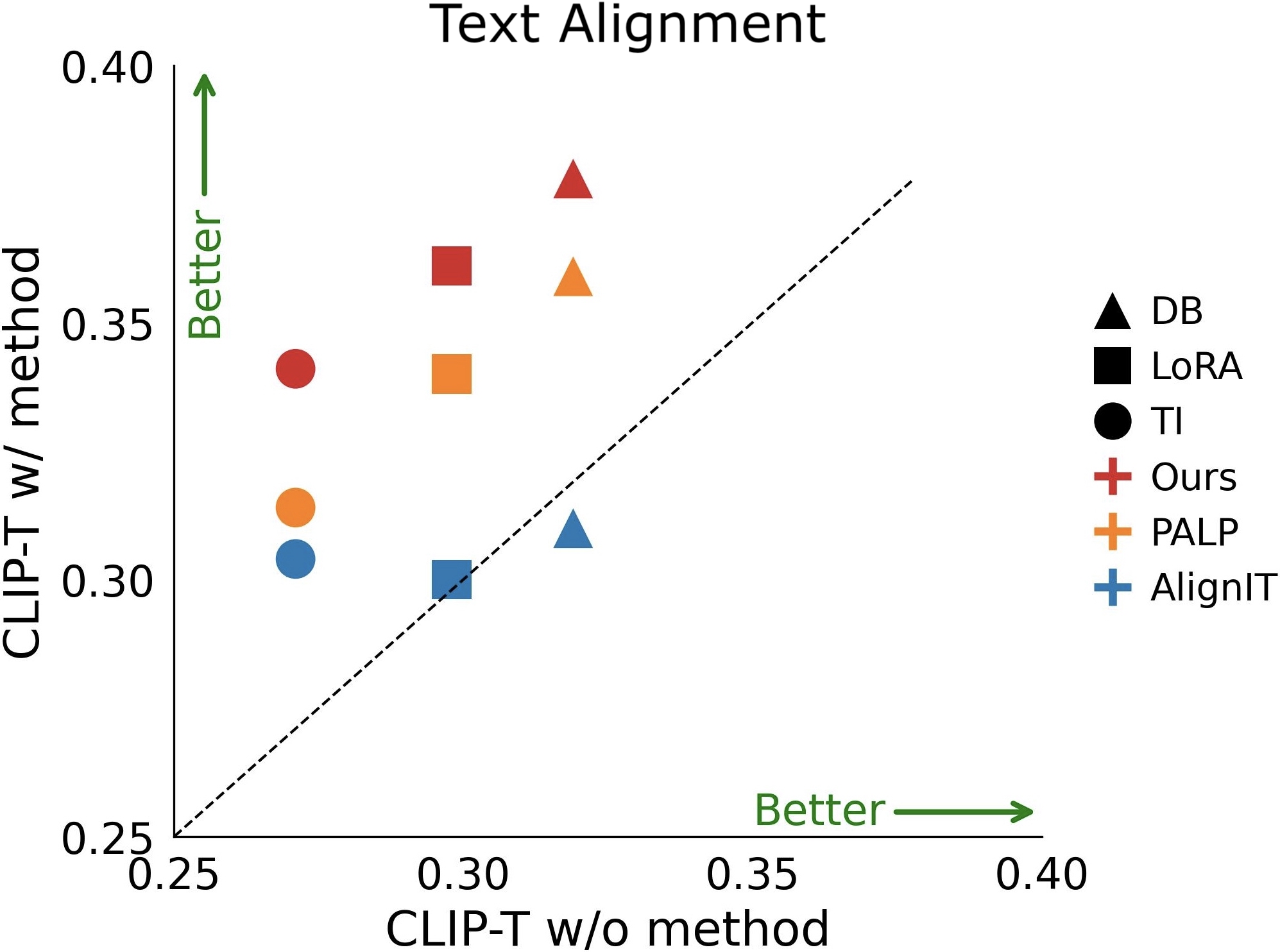}
    \caption{Quantitative Comparison with previous per-query methods. Image alignment (top) and text alignment (bottom) of various personalization approaches (including, $\text{DB }${\large$\blacktriangle$}, $\text{LoRA }\blacksquare$, and $\text{TI }${\large$\CIRCLE$}), with (w/) and without (w/o) the integration of different per-query methods (including, \textcolor{quan_blue}{\textbf{AlignIT}}, \textcolor{quan_orange}{\textbf{PALP}}, and \textcolor{quan_red}{\textbf{Ours}}).}
    \label{fig:qnc_vs_PQ}
\end{figure}

\subsubsection{Quantitative Evaluation}
Following previous work, we quantify the quality of our methods along two key dimensions: (1) Subject Alignment – how well the model preserves the identity of the subject, and (2) Prompt Adherence – how well the model aligns th generated image with the given prompt.

To estimate subject alignment, we use the evaluation benchmark, dataset, and metrics from DreamBooth~\cite{ruiz2023dreambooth}. We measure the average pairwise cosine similarity between ViT-S/16 CLIP~\cite{radford2021learning} embeddings of generated and real images (denoted as CLIP-I), as well as between ViT-S/16 DINO~\cite{caron2021emerging} embeddings (denoted as DINO-I). These metrics provide a proxy for how well the generated images maintain the identity of the original subject.

Prompt adherence is measured using CLIP-T, which calculates the average cosine similarity between the text prompt and the CLIP embedding of the generated image. This evaluates how accurately the generated image reflects the textual description.

Integrating our method into state-of-the-art baselines achieves superior performance on both subject alignment (CLIP-I) and prompt adherence (CLIP-T), as demonstrated in Fig.~\ref{fig:qnc_vs_SOTA} and \ref{fig:qnc_vs_PQ}.
\begin{figure}[b]
    \centering
    \includegraphics[width=1\linewidth]{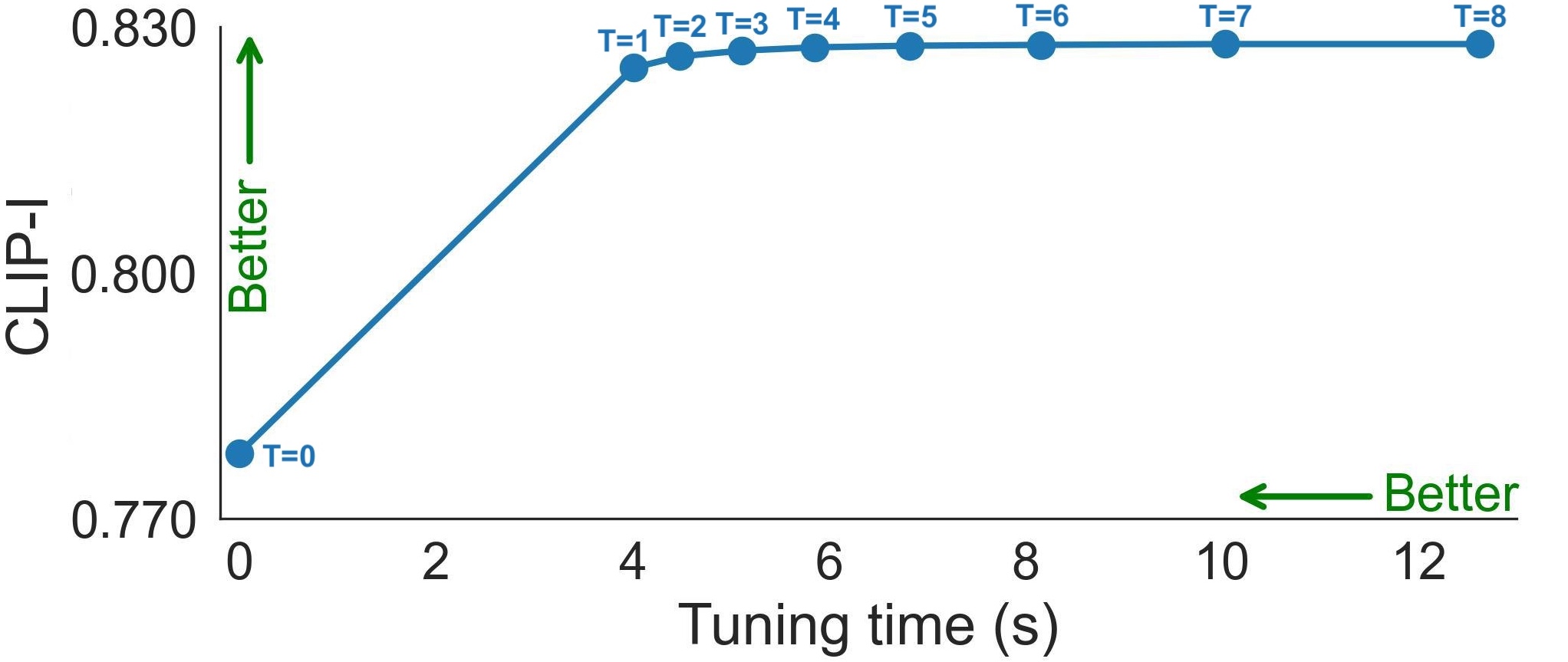}
    \hfill
    \includegraphics[width=1\linewidth]{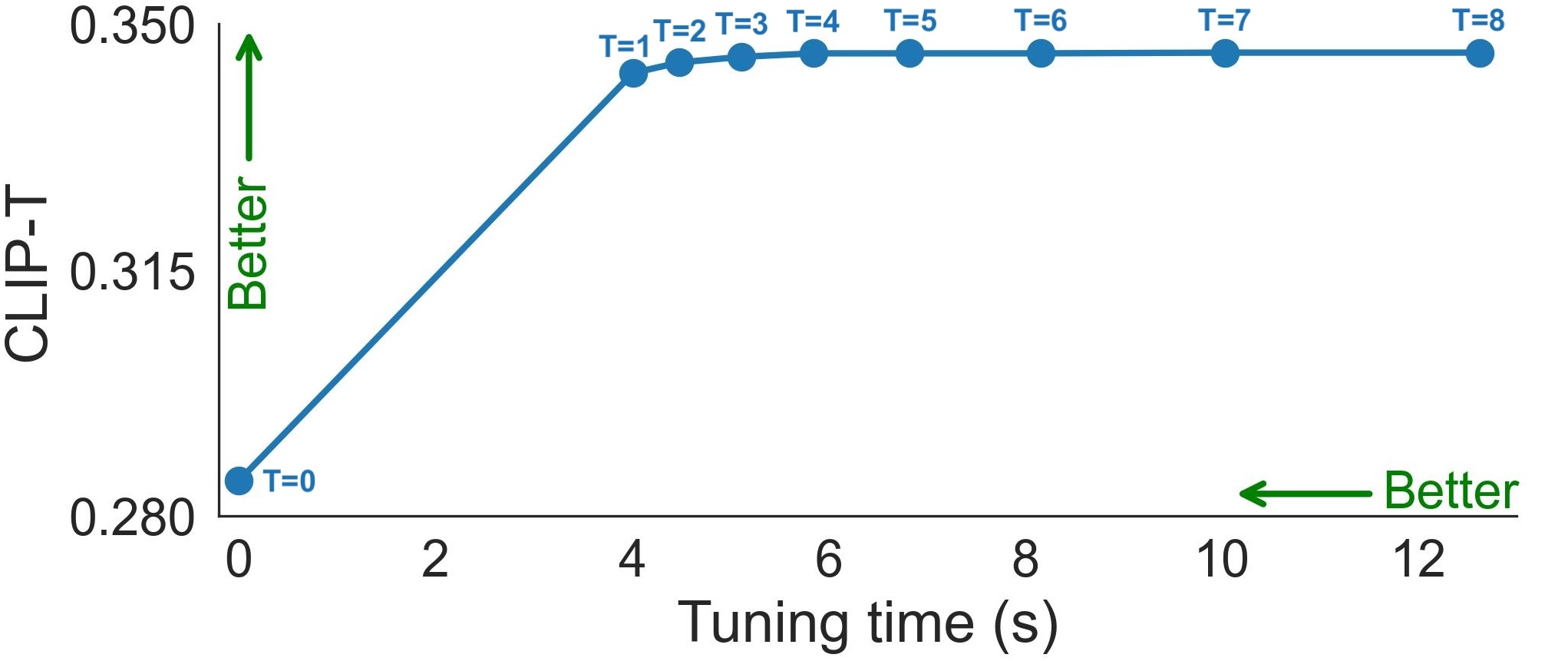}
    \caption{\textbf{Quality-vs-time tradeoff.} Figures show the CLIP-I (top) and CLIP-T (bottom) metrics as a function of fine-tuning duration. The duration is longer when using more features that are collected throughout the denoising path. $T$ is the number of feature maps using to compute the losses.}
    \label{fig:ts_time_opt1_ablation}
\end{figure}

\subsection{Ablation Studies and Sensitivity to Parameters}
\paragraph{Noise weight:} 
Figure~\ref{fig:noise_weight_ablation} shows using $t=1000$ when extracting the PDM features and cross-attention maps from $z_0$ for our loss produces the best performance.
\begin{table}[h]
    \centering
    \begin{minipage}{1\linewidth}
        \resizebox{1\linewidth}{!}{
            \begin{tabular}{c|c|c|ccc}
                \toprule
                \midrule
                \mc{3}{c}{\makecell{}}&
                \mc{2}{c}{\makecell{Subject Alignment}}&
                \mc{1}{c}{\makecell{Prompt\\Adherence}}\\
                \midrule
                \makecell{Method} &   
                \makecell{$\lambda_\text{PDM}$}\hspace{-2.5pt} &  
                \makecell{$\lambda_\text{CA}$}\hspace{-2.5pt} &  
                \makecell{DINO-I $\uparrow$}\hspace{-4pt} &  
                \makecell{CLIP-I $\uparrow$}\hspace{-4pt} &  
                \makecell{CLIP-T $\uparrow$}\hspace{-15pt}\\
                \midrule
                DB &  &  & $\textbf{0.659} \pm 0.101$ & $\textbf{0.805} \pm 0.048$ & $\textbf{0.296} \pm 0.023$\\
                DB & \checkmark &  & $0.631 \pm 0.104$ & $0.760 \pm 0.047$ & $0.272 \pm 0.024$\\
                DB &  & \checkmark & $0.660 \pm 0.101$ & $0.804 \pm 0.047$ & $0.300 \pm 0.024$\\

                \midrule

                TI &  &  & $\textbf{0.559} \pm 0.106$ & $\textbf{0.731} \pm 0.052$ & $\textbf{0.255} \pm 0.021$\\
                TI & \checkmark &  & $0.532 \pm 0.104$ & $0.711 \pm 0.056$ & $0.240 \pm 0.022$\\
                TI &  & \checkmark & $0.561 \pm 0.106$ & $0.730 \pm 0.052$ & $0.257 \pm 0.020$\\
                
                \bottomrule
            \end{tabular}
        }
    \vspace{-1.5mm}
    \caption{\textbf{Stage-wise loss effectiveness ablation study.} We examine the effectiveness of $\mathcal{L}_{\text{SA}}$ and $\mathcal{L}_{\text{CA}}$ throughout training rather than post-training, with two prominent methods, utilizing SD1.5 as backbone, with $\lambda_{\text{PDM}}=1,\lambda_{\text{CA}}=1$. We observe inferior results compared to post-training integration, and equal/inferior results compared to the baseline. This quantitatively support that certain losses become effective only after the model reaches a certain state.}
    \label{tab:stage_wise_ablation}
    \end{minipage}
\end{table}

\paragraph{Tuning-time:} We describe two sets of experiments designed to reduce the tuning time of our method.

First, shown in \cref{fig:ts_time_opt1_ablation}, we  calculated the loss across different stages of $z_t$. We found, calculating for the last stage (i.e., $z_0$) result in sufficient results, while collecting feature from other stages along the denoising path increase the tuning duration.

Second, shown in \cref{fig:ts_time_opt2_ablation}, we  looked into self-attention features from other layers in addition to the last layer used in PDM~\cite{samuel2024s}. The benefit of using more features was negligible, and did not justify  the increase in tuning time. using only the PDM features results in sufficient performance.

\paragraph{Stage-wise loss effectiveness:} Shown in \cref{tab:stage_wise_ablation}, we examined incorporating our proposed losses at other stages as well. Specifically, throughout training, across two prominent methods, using SD1.5~\cite{rombach2022high} as backbone. This ablation quantitatively supports that certain losses become effective only after the model reaches a certain state. Incorporating our objectives throughout training provides inferior results compared to the baseline, while using them post-training provides superior enhancement in terms of subject alignment and prompt adherence.
\begin{table}[h]
    \centering
    \begin{minipage}{1\linewidth}
        \resizebox{1\linewidth}{!}{
            \begin{tabular}{c|c|c|ccc}
                \toprule
                \midrule
                \mc{3}{c}{\makecell{}}&
                \mc{2}{c}{\makecell{Subject Alignment}}&
                \mc{1}{c}{\makecell{Prompt\\Adherence}}\\
                \midrule
                \makecell{Method} &   
                \makecell{$\lambda_\text{PDM}$}\hspace{-2.5pt} &  
                \makecell{$\lambda_\text{CA}$}\hspace{-2.5pt} &  
                \makecell{DINO-I $\uparrow$}\hspace{-4pt} &  
                \makecell{CLIP-I $\uparrow$}\hspace{-4pt} &  
                \makecell{CLIP-T $\uparrow$}\hspace{-15pt}\\
                \midrule
                DB & & & $0.659 \pm 0.101$ & $0.805 \pm 0.048$ & $0.296 \pm 0.023$\\
                DB &\checkmark& & $\textbf{0.709} \pm 0.104$ & $\textbf{0.844} \pm 0.045$ & $0.297 \pm 0.025$\\
                DB & &\checkmark& $0.658 \pm 0.103$ & $0.805 \pm 0.047$ & $\textbf{0.357} \pm 0.024$\\
                DB &\checkmark&\checkmark& $\textbf{0.710} \pm 0.103$ & $\textbf{0.845} \pm 0.046$ & $\textbf{0.358} \pm 0.025$\\

                \midrule

                TI & & & $0.559 \pm 0.106$ & $0.731 \pm 0.052$ & $0.255 \pm 0.021$\\
                TI &\checkmark& & $\textbf{0.644} \pm 0.100$ & $\textbf{0.809} \pm 0.047$ & $0.256 \pm 0.024$\\
                TI & &\checkmark& $0.558 \pm 0.107$ & $0.732 \pm 0.053$ & $\textbf{0.305} \pm 0.023$\\
                TI &\checkmark&\checkmark& $\textbf{0.646} \pm 0.101$ & $\textbf{0.810} \pm 0.048$ & $\textbf{0.306} \pm 0.022$\\
                
                \bottomrule
            \end{tabular}
        }
    \vspace{-1.5mm}
    \caption{\textbf{Per-loss contribution ablation study.} We investigate the contribution of each loss in our proposed method. Testing across different methods, leveraging SD1.5 as backbone, and using a fixed $\lambda_{\text{PDM}}=1,\lambda_{\text{CA}}=1$ values. $\mathcal{L}_{\text{PDM}}$ provides superior subject alignment, and $\mathcal{L}_{\text{CA}}$ results superior prompt-adherence. Incorporating both provides superior results in both aspects.}
    \label{tab:losses_contribution_ablation}
    \end{minipage}
\end{table}

\paragraph{Per-loss contribution:}
We quantify the contribution of each loss in our proposed method, by incorporating losses separately and together across different methods, using SD1.5~\cite{rombach2022high} as the backbone, while fixing $\lambda_{\text{SA}}=1,\lambda_{\text{CA}}=1$.
\cref{tab:losses_contribution_ablation} shows that  $\mathcal{L}_{\text{SA}}$ provided superior subject alignment, and $\mathcal{L}_{\text{CA}}$ resulted superior prompt-adherence. Incorporating both provided superior results in both aspects.

\ignore{ 
\section{Conclusion}
\label{sec:conclusion}
We introduced a fast per-query adaptation procedure that augments an already personalized text-to-image model to improve prompt adherence and identity preservation. By aligning attention-based identity features and prompt signals from the cross-attention maps at inference-time. The method improves both text alignment and visual consistency across multiple backbones and personalization methods, and handles complex prompts more reliably than prior per-query approaches. It is simple to adopt and strengthens personalization without retraining the full model. Limitations include reliance on a reasonably trained initial checkpoint and a current emphasis on single primary subjects. Future directions include multi-subject and interaction-aware refinement, as well as temporal extension to video.
}
\clearpage
\newpage
\clearpage
\newpage
\section{Acknowledgements}
This study was funded by the Israeli Ministry of Science, Israel-Singapore binational grant, and by a grant from the Israeli higher-council of education, through the Bar-Ilan data science institute (BIU DSI).

{\small
    \bibliographystyle{ieeenat_fullname}
    \bibliography{main}
}
\clearpage
\appendix

\section{Additional qualitative results}
\begin{figure*}[hb!]
  \centering
  \includegraphics[width=1\linewidth]{imgs/qlc/vs_SOTA/models.png}
  \includegraphics[width=1\linewidth]{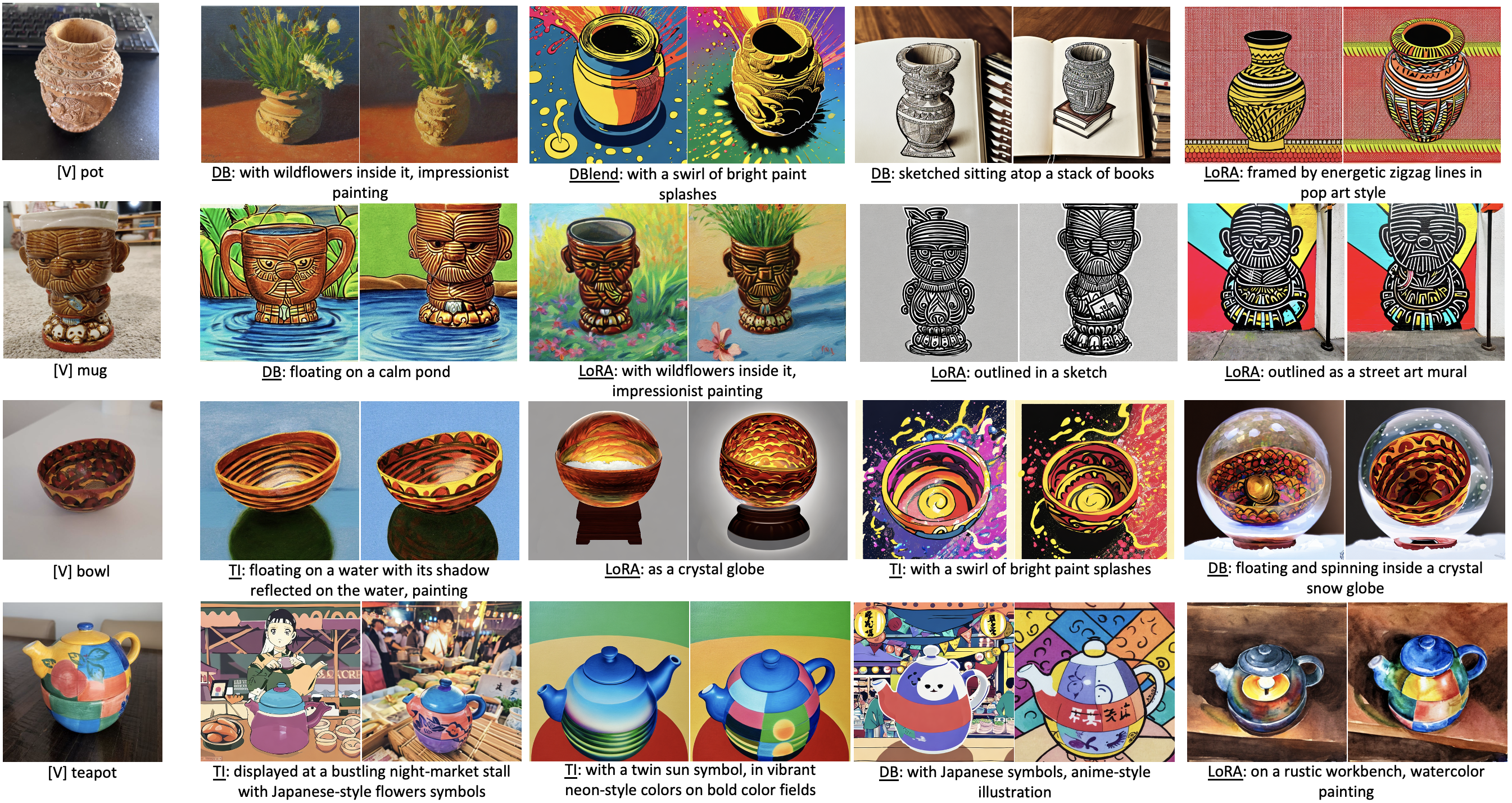}
  \caption{\textbf{Additional qualitative comparison}. We present images generated by various personalization methods without our method and with it, including DB~\cite{ruiz2023dreambooth}, TI~\cite{gal2022image}, DBlend~\cite{ram2025dreamblend}, CLD~\cite{huang2024classdiffusion}, AttnDB~\cite{pang2024attndreambooth}, and LoRA~\cite{LoRADiffusion}, across 4 different backbones (FLUX~\cite{labs2025flux1kontextflowmatching}, SDXL~\cite{podell2023sdxl}, SD2.1, SD1.5~\cite{rombach2022high}). Adding our method to those personalization approaches demonstrates superior performance in text alignment and identity preservation compared to these baselines.}
  \label{fig:qlc_vs_SOTA_more}
\end{figure*}

\begin{figure*}[hb!]
  \centering
  \includegraphics[width=1\linewidth]{imgs/qlc/vs_PQ/methods.png}
  \includegraphics[width=1\linewidth]{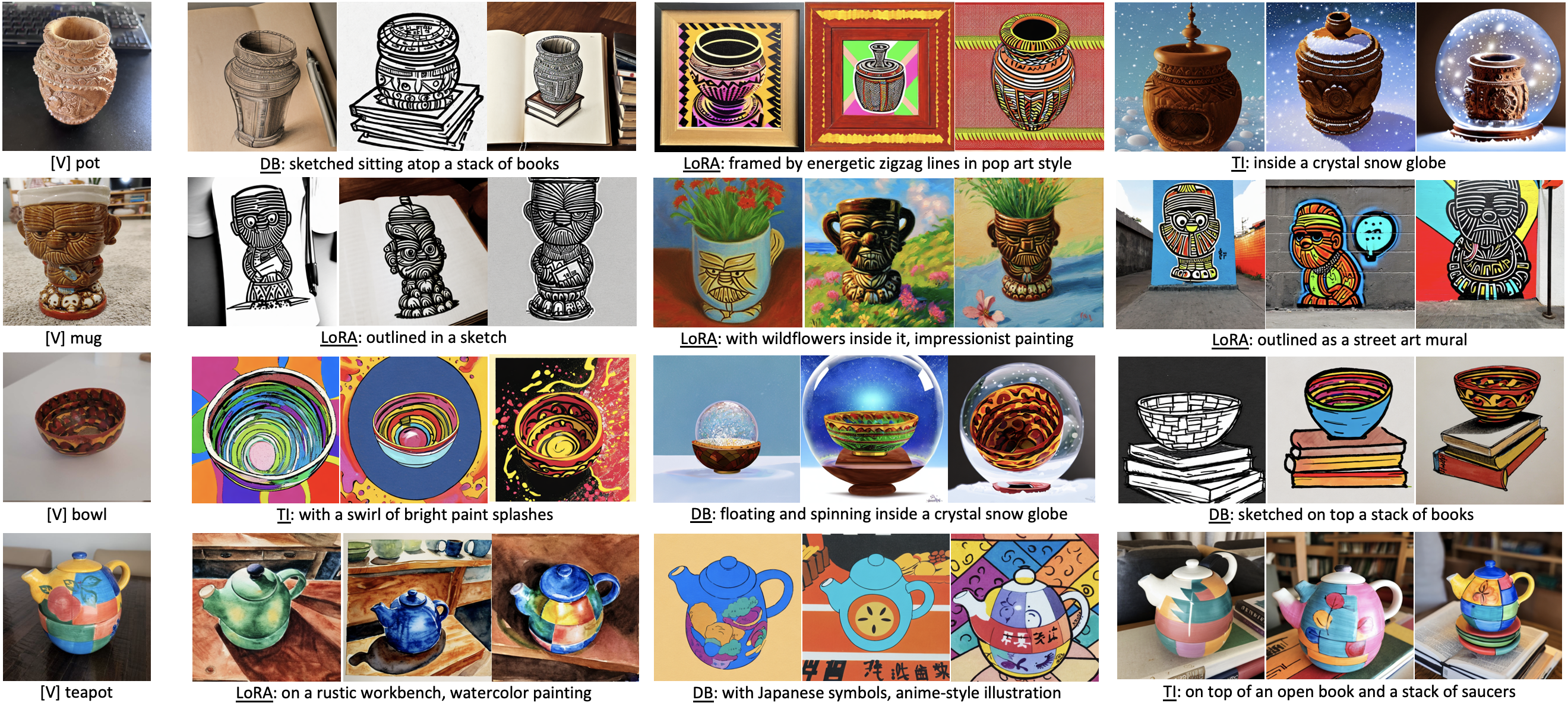}
  \includegraphics[width=1\linewidth]{imgs/qlc/vs_PQ/models.png}
  \caption{\textbf{Additional qualitative comparison}. We present images generated by 3 personalization methods (including DB~\cite{ruiz2023dreambooth}, TI~\cite{gal2022image}, LoRA~\cite{LoRADiffusion}) using 3 per-query methods (including AlignIT~\cite{agarwal2025alignit}, PALP~\cite{arar2024palp}, Ours), across 3 different backbones (FLUX~\cite{labs2025flux1kontextflowmatching}, SDXL~\cite{podell2023sdxl}, SD2.1~\cite{rombach2022high}). Our method demonstrates superior performance in text alignment and identity preservation compared to these per-query methods.}
  \label{fig:qlc_vs_PQ_more}
\end{figure*}

We present additional qualitative comparisons, showcasing results with and without our method applied to various personalization techniques (Figure \ref{fig:qlc_vs_SOTA_more}), as well as comparisons against other per-query methods (Figure \ref{fig:qlc_vs_PQ_more}). These figures highlight that our approach significantly enhances generation quality by improving both image alignment (better visual consistency with the reference image) and text alignment (more faithful adherence to the prompt).

\section{Ablation Study}
\begin{figure}[h!]
    \centering 
    \includegraphics[width=0.88\linewidth]{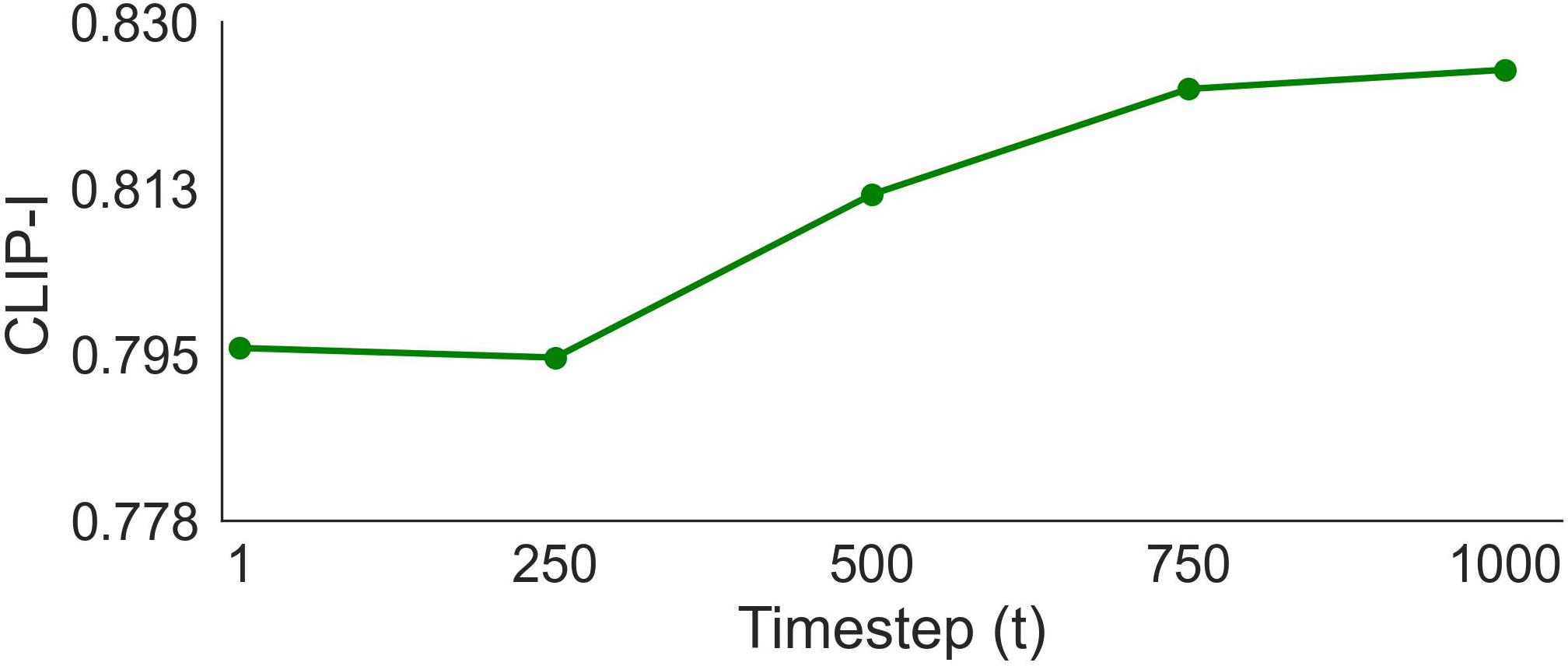}
    \includegraphics[width=0.88\linewidth]{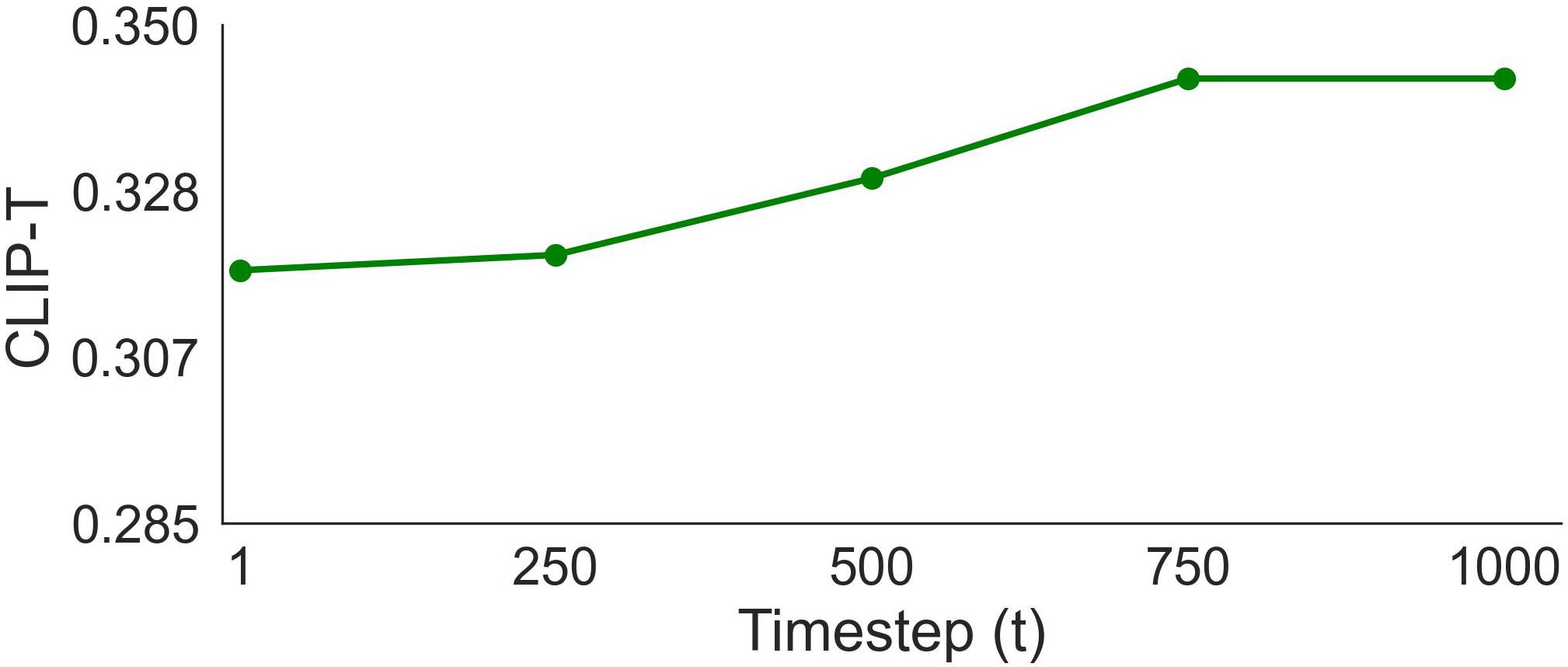}
    \caption{\textbf{Noise weight Ablation Study.} Sensitivity of CLIP-I and CLIP-T metrics to the magnitude of the noise, in terms of the $t$ parameter used when calculating features.}
    \label{fig:noise_weight_ablation}
\end{figure}
Section~\ref{sec:method} describes the feature extraction process. In the main paper, we use a fixed value of $t=1000$. Here, we explore how varying $t$ affects the results. Figure \ref{fig:noise_weight_ablation} presents CLIP-I and CLIP-T scores across different $t$ values on the DreamBooth dataset. The results indicate that $t=1000$ consistently yields the best performance.

\begin{figure}[h!]
    \centering
    \includegraphics[width=1\linewidth]{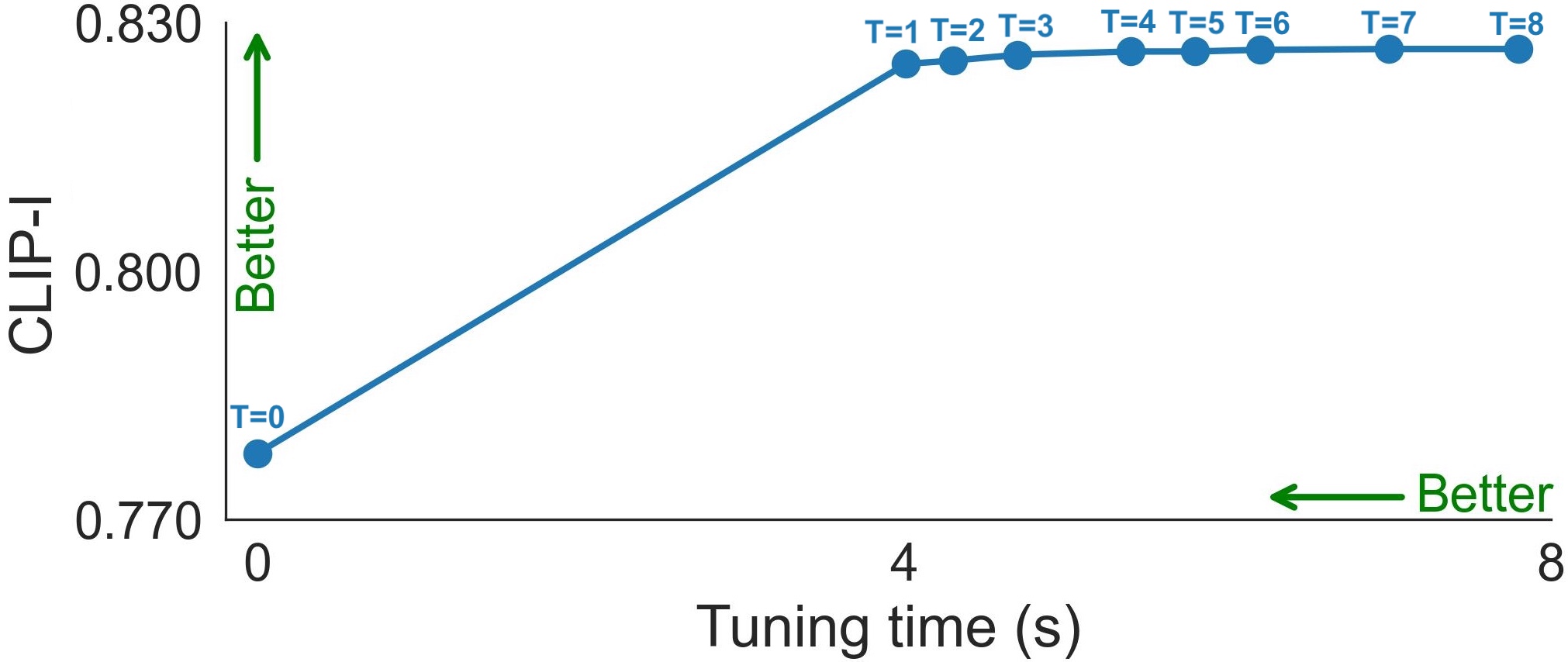}
    \hfill
    \includegraphics[width=1\linewidth]{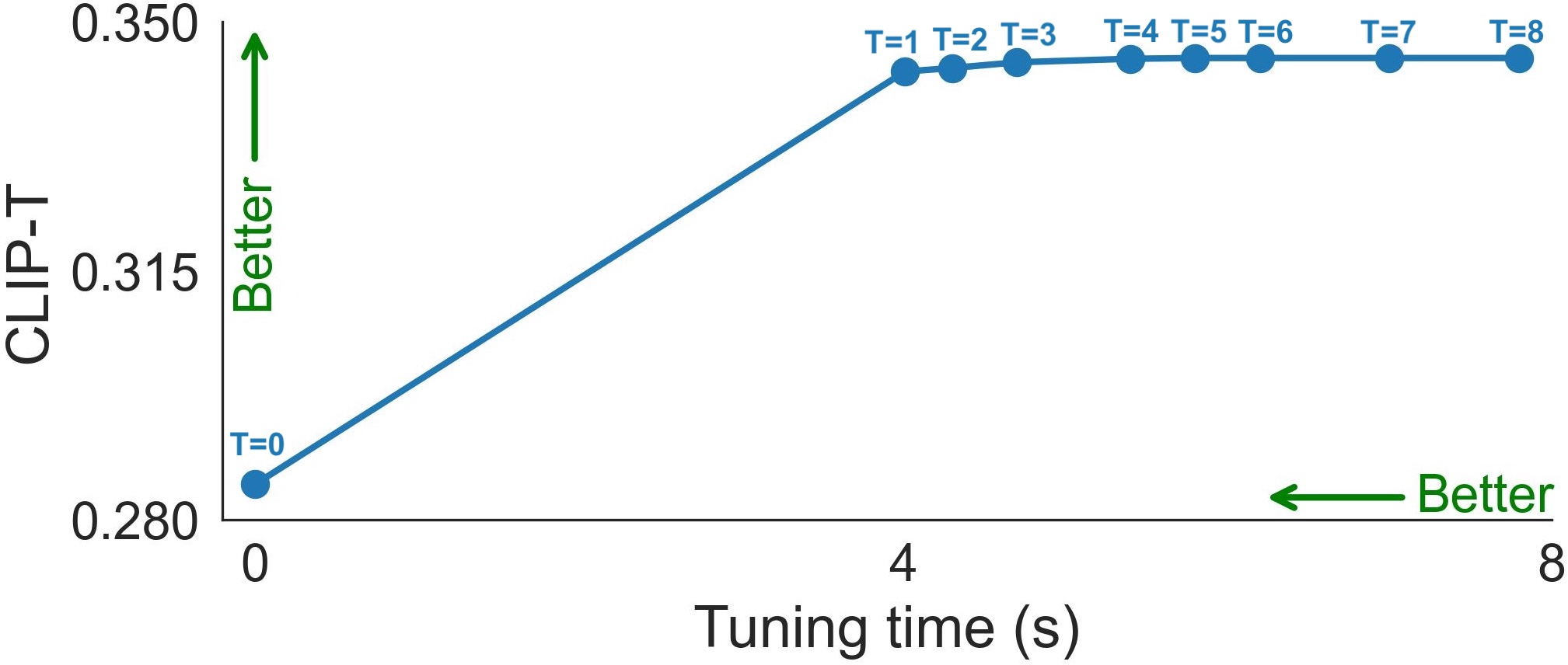}
    \caption{\textbf{Quality-vs-time tradeoff.} Figures show the CLIP-I (top) and CLIP-T (bottom) metrics as a function of fine-tuning duration. The duration is longer when using more SA features that are collected throughout the denoising path. $T$ is the number of feature maps using to compute the losses.}
    \label{fig:ts_time_opt2_ablation}
\end{figure}

\newpage

\end{document}